\providecommand{\main}{./Results}

\documentclass[%
onecolumn
, hidempi
]{mpi2015-cscpreprint}
\makeatletter
\DeclareOldFontCommand{\rm}{\normalfont\rmfamily}{\mathrm}
\DeclareOldFontCommand{\sf}{\normalfont\sffamily}{\mathsf}
\DeclareOldFontCommand{\tt}{\normalfont\ttfamily}{\mathtt}
\DeclareOldFontCommand{\bf}{\normalfont\bfseries}{\mathbf}
\DeclareOldFontCommand{\it}{\normalfont\itshape}{\mathit}
\DeclareOldFontCommand{\sl}{\normalfont\slshape}{\@nomath\sl}
\DeclareOldFontCommand{\sc}{\normalfont\scshape}{\@nomath\sc}
\makeatother

\usepackage{hyperref}       
\usepackage{url}            
\usepackage{booktabs}       
\usepackage{amsfonts}       
\usepackage{nicefrac}       
\usepackage{microtype}      
\usepackage{xcolor}         

\usepackage{graphicx}
\usepackage{caption}
\usepackage{subcaption}

\usepackage{amssymb}
\usepackage{amsthm}
\usepackage[mathscr]{eucal}

\usepackage[american]{babel}
\usepackage{multirow}

\usepackage[software, hardware]{mymacros}

\usepackage{cleveref}
\usepackage{grffile}

\renewcommand{\d}{\texttt{{d}}}
\newcommand{\rk}{{\texttt{RK4}}}

\newcommand{\dt}{\texttt{{dt}}}

\newcommand{\inode}{\texttt{Imp-NODEs}}
\newcommand{\snode}{\texttt{Std-NODEs}}
\newcommand{\mse}{\texttt{MSE}}
\renewcommand{\grad}{\texttt{Grad}}
\newcommand{\integral}{\texttt{Integral}}

\usepackage{todonotes}
\usepackage{sidecap}

\begin{document}

\title{Neural ODEs with Irregular and Noisy Data}

\author[$\ast$]{Pawan Goyal}
\affil[$\ast$]{Max Planck Institute for Dynamics of Complex Technical Systems, 39106 Magdeburg, Germany.\authorcr
	\email{goyalp@mpi-magdeburg.mpg.de}, \orcid{0000-0003-3072-7780}
}

\author[$\dagger$]{Peter Benner}
\affil[$\dagger$]{Max Planck Institute for Dynamics of Complex Technical Systems, 39106 Magdeburg, Germany.\authorcr
	\email{benner@mpi-magdeburg.mpg.de}, \orcid{0000-0003-3362-4103}
}

\shorttitle{Neural ODEs with Irregular and Noisy Data}
\shortauthor{P. Goyal, P. Benner}
\shortdate{}

\code{\url{https://gitlab.mpi-magdeburg.mpg.de/goyalp/implicit_neuralodes}}

\abstract{%
Measurement noise is an integral part while collecting data of a physical process. Thus, noise removal is necessary to draw conclusions from these data, and it often becomes essential to construct dynamical models using these data.
We discuss a methodology to learn differential equation(s) using noisy and irregular sampled measurements. In our methodology, the main innovation can be seen in the integration of deep neural networks with the neural ordinary differential equations (ODEs) approach.
Precisely, we aim at learning a neural network that provides (approximately) an implicit representation of the data and an additional neural network that models the vector fields of the dependent variables. We combine these two networks by constraining using neural ODEs. 
The proposed framework to learn a model describing the vector field is highly effective under noisy measurements. The approach can handle scenarios where dependent variables are not available at the same temporal grid. 
Moreover, a particular structure, e.g., second-order with respect to time, can easily be incorporated.
We demonstrate the effectiveness of the proposed method for learning models using data obtained from various differential equations and present a comparison with the neural ODE method that does not make any special treatment to noise.
%
}

\novelty{
	\begin{itemize}
		\item This work blends neural networks with the neural ODE methods to learn dynamical models from noisy measurements, as well as to obtain denoised data. 
		\item More precisely, two networks, one for an implicit representation of data and the second one for describing dynamics, are combined by an integral form of ODEs. 
		\item An extension to second-order ODEs is discussed. 
		\item The proposed methodology is capable of a situation when the dependent variables are sampled irregularly. Moreover, they need not be measured on the same irregular time grid.
	\end{itemize} 
}
\maketitle

\section{Introduction}\label{sec:introduction}
Uncovering dynamical models explaining physical phenomena and dynamic behaviors has been active research for centuries \footnote{For example, Isaac Newton developed his fundamental laws based on measured data.}. When a model describing the underlying dynamics is available, it can be used for several engineering studies such as process design, optimization, predictions, and control. Conventional approaches based on physical laws and empirical knowledge are often used to derive dynamical models. However, this is impenetrable for many complex systems,  e.g., understanding the Arctic ice pack dynamics, sea ice, power grids, neuroscience, or finance, to only name a few applications. Data-driven methods to discover models have enormous potential to understand transient behaviors in the latter cases better. Furthermore, data acquired using imaging devices or sensors are contaminated with measurement noise. Therefore, systematic approaches that learn a dynamic model with proper treatment of noise are required.  


Towards this aim, the initial work~\cite{rudy2019deep} proposes a framework that explicitly incorporates the noise into a numerical time-stepping method, namely a \emph{Runge-Kutta} method. Though the approach has shown promising directions, its scalability remains ambiguous as the approach explicitly needs noise estimates and aims to decompose the signal explicitly into noise and ground truth. Moreover, it requires that the Runge-Kutta method can give a reasonable estimate at the next step. Additionally, irregular sampling (e.g., when dependent variables are not even collected at the same time-grid) cannot be applied, which can be highly relevant when information is gathered from various sources, e.g., medical applications. This work discusses a deep learning-based approach to learning a dynamic model by attenuating neural networks with adaptive numerical integrations. It allows learning models to represent the vector field accurately without estimating noise explicitly and when dependent variables are arbitrarily irregularly sampled. 

\begin{figure}[!tb]
	\centering
	\includegraphics[width= \textwidth]{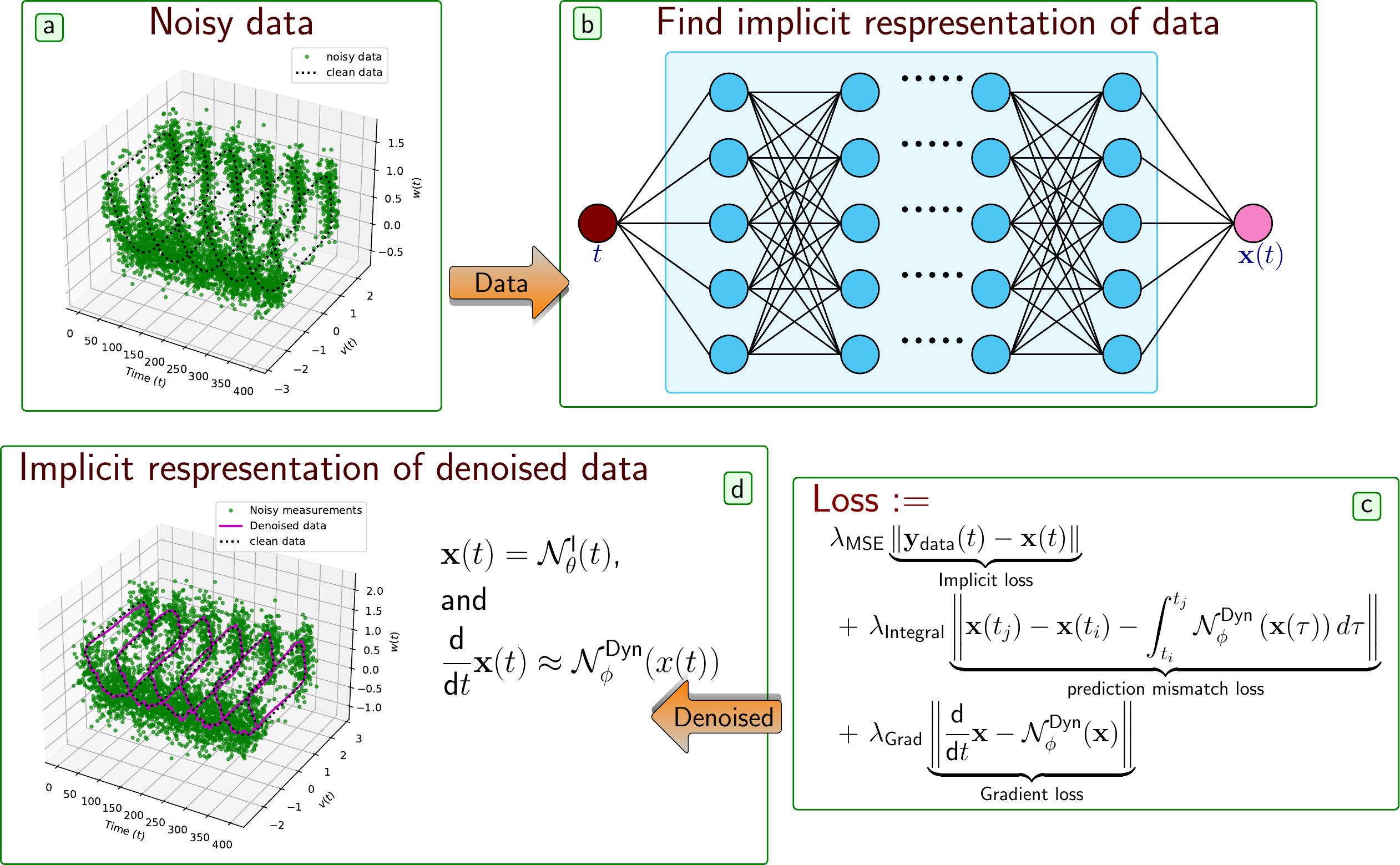}
	\caption{The figure illustrates the framework for de-noise data and learning a model describing underlying dynamics. For this, we determine an implicit representation of data (approximately) by a network $\cN^{\textsf{I}}_\theta$ and another network for the vector field  $\cN^{\textsf{Dyn}}_\phi$. These two networks are connected by enforcing that the dynamics of the output of the implicit representation can be given by $\cN^{\textsf{Dyn}}_\phi$. Once the loss is minimized (shown in (c)), we obtain an implicit network for de-noised data and a model for the vector field $\cN^{\textsf{Dyn}}_\phi$.}
	\label{fig:method_overview}
\end{figure}

\paragraph{Our contributions:} 
Our work introduces a framework to learn dynamics models by innovatively blending neural networks and numerical integration methods from noisy and irregular measurements. Precisely, we aim at learning two networks; one that approximately represents given measurement data implicitly, and the second one approximates the vector field. We connect these two networks by enforcing an integral form of ODEs as depicted in \Cref{fig:method_overview}. The appeal of the approach is that we do not require an explicit estimate of noise to learn a model. Furthermore, the proposed approach is applicable even if each dependent variable is collected on a different time grid, which can be irregular. 

The remaining structure of the paper is as follows. In the next section, we present a summary of relevant work. In \Cref{sec:method}, we present our deep learning-based framework for learning dynamics from noisy measurements by combining two networks. There, we also demonstrate the effectiveness of the proposed methodology using various synthetic data with increasing noise levels. \Cref{sec:secondOrder} discusses the application of learning second-order dynamical models. Moreover, in \Cref{sec:irregularsampling}, we discuss how to handle irregular sampling of measurements. We conclude the paper with a summary and future research directions. 
\section{Relevant work}\label{sec:relaventwork}
Data-driven methods to learn dynamic models have been studied for several decades, see, e.g., \cite{juang1994applied,ljung1999system,billings2013nonlinear}. Learning linear models from input-output data goes back to Ho and Kalman \cite{ho1966effective}. There have been several algorithmic developments for linear systems, for example, the eigensystem realization algorithm  \cite{juang1985eigensystem,longman1989recursive}, and Kalman filter-based approaches \cite{juang1993identification,phan1993linear,phan1992identification}. Dynamic mode decomposition has also emerged as a promising approach to construct models from input-output data and has been widely applied in fluid dynamics applications, see, e.g., \cite{kalman1960new, schmid2010dynamic,tu2014dynamic}. Furthermore, there has been a series of developments to learn nonlinear dynamic models. This includes, for example,  equations free modeling \cite{kevrekidis2003equation}, nonlinear regression \cite{voss1999amplitude}, dynamic modeling \cite{ye2015equation}, and automated inference of dynamics \cite{schmidt2011automated,daniels2015automated, daniels2015efficient}. Utilizing symbolic regression and an evolutionary algorithm \cite{bongard2007automated, schmidt2009distilling}, learning compact nonlinear models becomes possible. Moreover, leveraging sparsity (also known as sparse regression), several approaches have been proposed \cite{brunton2016sparse,mangan2016inferring,tran2017exact, schaeffer2020extracting, mangan2017model, morGoyB21a}. We also mention the work \cite{raissi2018hidden} that learns models using Gaussian process regression. All these methods have particular approaches to handle noise in the data. For example,  sparse regression methods, e.g., \cite{brunton2016sparse, mangan2016inferring, morGoyB21a}  often utilize smoothing methods before identifying models, and the work \cite{raissi2018hidden} handles measurement noise as data represented like a Gaussian process. 

Even though the aforementioned nonlinear modeling methods are appealing and powerful in providing analytic expressions for models, they are often built upon model hypotheses. For example, the success of sparse regression techniques relies on the fact that the nonlinear basis functions, describing the dynamics, lie in a candidate features library. For many complex dynamics, such as the melting Arctic ice, the utilization of these methods is not trivial. Thus, machine learning techniques, particularly deep learning-based ones, have emerged as powerful methods capable of expressing any complex function in a black-box manner given enough training data. Neural network-based approaches in the context of dynamical systems have been discussed in \cite{chen1990non,rico1993continuous,gonzalez1998identification,milano2002neural} decades ago. A particular type of neural network, namely recurrent neural networks, intrinsically models sequences and is often used for forecasting \cite{lu2018attractor,pan2018long,pathak2017using,pathak2018hybrid,vlachas2018data} but does not explicitly learn the corresponding vector field. Deep learning is also utilized to identify a coordinate transformation so that the dynamics in the transformed coordinates are almost linear or sparse in a high-dimensional feature basis, see, e.g., \cite{lusch2018deep, takeishi2017learning, yeung2019learning, champion2019data}. Furthermore, we mention that classical numerical schemes are incorporated with feed-forward neural networks to have discrete-time steppers for predictions, see \cite{gonzalez1998identification,raissi2018multistep,raissi2019physics,raissi2020hidden}. The approaches in \cite{gonzalez1998identification, raissi2018multistep} can be interpreted as nonlinear autoregressive models~\cite{billings2013nonlinear}.  
A crucial feature of deep learning-based approaches that integrates numerical integration schemes is that vector fields are estimated using neural networks. Also, time-stepping is done using a numerical integration scheme. Furthermore, in recent times, \emph{neural ordinary differential equations} (Neural ODEs) in which neural networks define the vector fields, have been proposed in \cite{chen2018neural}, where it is shown how to compute gradient with respect to network parameters efficiently using adjoint sensitivities. As a result, one can utilize efficient black-box numerical solvers to solve ODEs in a given time span using adaptive time-stepping. 

However, measurement data are often corrupted with noise, and these mentioned approaches do not perform any specific noise treatment. The work in \cite{rudy2019deep} proposes a framework that explicitly incorporates the noise into a numerical time-stepping method. Though the approach has shown a promising direction, its scalability remains ambiguous. The approach explicitly needs noise estimates by learning the decomposition of the signal into noise and ground truth. Also, it relies on a Runge-Kutta scheme that can accurately estimate the variable at the next step. 

Furthermore, in scenarios where the data are collected on an irregular time grid, the work \cite{rubanova2019latent} discussed a methodology by combining gated recurrent unit (GRU) and neural ODEs. In the approach, an estimate for the initial condition of (latent) ODEs is learned, and an ODE for the vector field is then integrated using the estimated initial condition. However, long sequences are quite challenging to estimate the initial condition given measurements future in time. Although in \cite{rubanova2019latent}, the measurements can be collected at an irregular time grid, it still requires that all dependent variables are measured at the same time grid. In the cases when each dependent variable is measured at a different time-grid, then the approach \cite{rubanova2019latent} is not applicable. 


\section{Proposed Methodology---\textbf{Implicit Networks Combined Neural ODEs}}\label{sec:method}
In this section, we discuss our framework to learn dynamical models using noisy measurements without explicitly estimating noise. To achieve the goal, we utilize the powerful approximation capabilities of deep neural networks and their automatic differentiation feature with the neural ODEs approach \cite{chen2018neural}, which allows integrating a function, defining the vector field,  with any desired method and accuracy. 
For this, let us consider an autonomous nonlinear differential equation:
\begin{equation}
	\dfrac{\d}{\d t}\bx(t) = \bg(\bx(t)), \quad \bx(0) = \bx_0,
\end{equation}
where $\bx(t)\in \R^{n}$ denotes the solution at time $t$, and the continuous function $\bg(\cdot): \Rn \rightarrow \Rn$ defines the vector field. Furthermore, the solution $\bx(t_{j})$ can be explicitly given as:
\begin{equation}\label{eq:connect_xi_xj}
	\bx(t_{j}) = \bx(t_{i}) + \int_{t_i}^{t_{j}}\bg(\bx(\tau)) \d\tau. 
\end{equation}

Next, we further proceed to discuss our framework to learn dynamical models from noisy measurements. The approach involves two networks.  The first network implicitly represents the variable as shown in \Cref{fig:method_overview}(b), and the second network approximates the vector field, or the function $\bg(\cdot)$. These two networks are related by connecting the dependent variables at time $t_i$ and $t_j$ as shown in \eqref{eq:connect_xi_xj}. That is, the output of the implicit network is not only in the vicinity of the measurement data but also its time-evolution can be defined by $\bg(\bx)$ or \eqref{eq:connect_xi_xj}. To be mathematically precise, let us denote noisy measurement data at time $t_i$ by $\by(t_i)$. Furthermore, we consider a feed-forward neural network, denoted by 
$\cN^{\textsf{I}}_{\theta}$ parameterized by $\theta$, that approximately yields an implicit representation of measurement data, i.e., 
\begin{equation}
	\cN^{\textsf{I}}_{\theta}(t_i) := \bx(t_i)    \approx \by(t_i),
\end{equation}
where $i \in \{1,\ldots, m\}$ with $m$ being the total number of measurements.  Additionally, let us denote another neural network by $\cN^{\textsf{Dyn}}_{\phi}$ parameterized by $\phi$ that approximates the vector field $\bg(\cdot)$. We connect these two networks by enforcing that the time-evolution of the output of the network $\cN^{\textsf{I}}_{\theta}$ can be described by $\cN^{\textsf{Dyn}}_{\phi}$, i.e., 
\begin{equation}
	\bx(t_{i+1})  \approx \bx(t_{i}) + \int_{t_i}^{t_j} \bg(\bx(\tau))d\tau, \quad \text{and}\quad
	\dfrac{\d}{\d t}  \bx(t_i) \approx \cN^{\textsf{Dyn}}_{\phi}\left(\bx(t_i)\right).
\end{equation}
As a result, our goal becomes to determine the network parameters $\{\theta, \phi\}$ such that the following loss is minimized:
\begin{equation}\label{eq:loss_func}
	\cL = \lambda_{\mse} \cdot \cL_{\mse} +  \lambda_{\integral} \cdot \cL_{\integral}  + \lambda_{\grad} \cdot \cL_{\grad}, 
\end{equation}
where 
\begin{itemize}
	\item  $\cL_{\mse} $ denotes the mean square error of the output of the network $\cN^{\textsf{I}}_{\theta}$ and noisy measurements, i.e., 
	\begin{equation}
		\left\|\cN^{\textsf{I}}_{\theta}(t_i) -  \by(t_i)\right\|^2_{F}.
	\end{equation}
	The loss enforces measurement data to in the vicinity of the output of the implicit network, and $ \lambda_{\mse}$ is its weighting parameter. 
	\item The term $\cL_{\integral}$ links the two networks by comparing the prediction, i.e.,
	\begin{equation}
		\left\|\bx(t_{j}) - \bx(t_{i}) - \int_{t_i}^{t_j}\bg(\bx(\tau))d\tau\right\|^2_F,
	\end{equation}
	and the parameter $\lambda_{\integral}$ defines its weight  in the total loss.
	\item The vector field at the output of the implicit network can also be computed directly using automatic differentiation, but it also can be computed using the network~$\cN^\textsf{Dyn}_\phi$. The term $\cL_{\textsf{Grad}}$ penalizes its mismatch as follows:
	\begin{equation}
		\left\|\cN_\phi^{\textsf{Dyn}}(\bx (t_i))  - \dfrac{\d}{\d t} \bx(t_i) \right\|^2_F,
	\end{equation}
	and $\lambda_{\grad}$ is its corresponding regularization parameter. 
\end{itemize}
The total loss $\cL$ can be minimized using a gradient-based optimizer such as Adam \cite{kingma2014adam}. Once the networks are trained and have found their parameters that minimize the loss, we can generate the denoised variables using the implicit network $\cN_\theta^{\textsf{I}}$, and the vector field by the network $\cN_\phi^{\textsf{Dyn}}$. In the rest of the paper, we denote the proposed methodology using Implicit--Neural ODEs (in sort \inode). 
\section{Numerical Experiments}
We now present the performance of the approach discussed in \Cref{sec:method} to de-noise measurement data, as well as to learn a model for estimating the vector field by means of an example. To that aim, we consider data obtained by solving a differential equation that is then corrupted using white Gaussian noise by varying the noise level.   For a given percentage, we determine the noise as follows:
\begin{equation*}
	\nu \sim \cN\left(0,\mu\right), \quad \text{with}~~ \mu = \mathsf{Noise}\%.
\end{equation*}
We have implemented our framework using the deep learning library \texttt{PyTorch} \cite{paszke2019pytorch} and have optimized both networks together using the Adam optimizer \cite{kingma2014adam}. We have used \texttt{torchdiffeq} \cite{chen2018neural}, a Python package to integrate an ODE and to do back-propagation to determine gradients, with the default settings. 

Moreover, to obtain a good initial guess to employ the proposed \inode, we replace the integration term by its approximation using $4^{\textsf{th}}$-order Runge-Kutta (\rk) method. It is computationally faster because the \rk~method takes the fixed four calls of the function, defining the vector field. We first train using this strategy for $5~000$ epochs, followed by training using an adaptive ODE integration scheme within a given time span for $10~000$ epochs for $\mu = \{1\%, 5\%\}$, and for $1000$ epochs only for $\mu = \{10\%,\ldots,50\%\}$ to do early stopping to avoid over-fitting. We also make use of a learning scheduler, for which we reduced the learning rate by one-tenth after each $4~000$ epoches. The neural network architecture design and hyperparameters are discussed in \Cref{appendix:models}, and we have run all our experiments on a \nvidia~\texttt{P100 GPU}.

\paragraph{Cubic damped model:}We consider a damped cubic system, which is described by
\begin{equation}
	\begin{aligned}
		\dot\bx(t) &= -0.1\bx(t)^3 + 2.0\by(t)^3,\\
		\dot\by(t) &= -2.0\bx(t)^3 - 0.1\by(t)^3.
	\end{aligned}
\end{equation}
It has been one of the benchmark examples in discovering models using data, see, e.g., \cite{brunton2016discovering,morGoyB21a} but there, it is assumed that the dynamics can be given sparsely in a high-dimensional feature dictionary. Here, we do not make any such assumptions and instead learn the vector field using a neural network. For this example, we take $2~500$ data points in the time interval $[0,25]$ by simulating the model using the initial condition $[2,0]$. We add various noise levels in the clean data to have noisy measurements synthetically. We construct neural networks for the implicit representation and the vector field with the parameters given in \Cref{tab:NN_info_ODE}.  

We corrupt the data by adding mean-zero Gaussian white noise of variances $\{1\%,\ldots, 50\%\}$. We aim to obtain a denoised signal and a model, defining its vector field. Before employing the method, we perform a pre-processing step to noisy data using a low-pass filter to remove a large portion of the high-frequency noise. 
We compare our methodology with the neural ODE framework \cite{chen2018neural}, which also focuses on learning a neural network, defining the underlying vector field. 

To train the  implicit network, and the neural network for ODEs, we set $\lambda_{\integral} = 1.0$ and $\lambda_{\textsf{Grad}} = 10^{-2}$ in the loss function \eqref{eq:loss_func}, and choose $\lambda_{\mse} = 1.0$ for $\mu = \{1\%, 5\%\}$, and $\lambda_{\mse} = 0.5$ for  $\mu = \{10\%, 20\%\}$, and $\lambda_{\mse} = 0.2$ for $\{30\%, 40\%, 50\%\}$ to avoid over-fitting of noisy data using the implicit network. Moreover, to integrate  ODEs, we consider the time-span of $\texttt{bs}\cdot \dt$, where $\texttt{bs}$ is treated as a batch-size and is set to $4$ and $\dt = 10^{-2}$. 

\begin{figure}[!tb]
	\begin{tikzpicture}[scale=0.7, every node/.style={scale=0.7}]
		\node[] (l1) {};
		\node[right = -0.0cm of l1, text width = 0.25\textwidth] (l2) {True vector field (VF)};
		\node[right = 0.35cm of l2] (l3) {\inode};
		\node[right = 1.45cm of l3] (l4) {Errors b/w VFs};
		\node[right = 1.20cm of l4] (l5) {\snode~\cite{chen2018neural}};
		\node[right = 1.45cm of l5] (l6) {Errors b/w VFs};
	\end{tikzpicture}
	\includegraphics[width = 0.57\textwidth,height = 2.2cm, ]{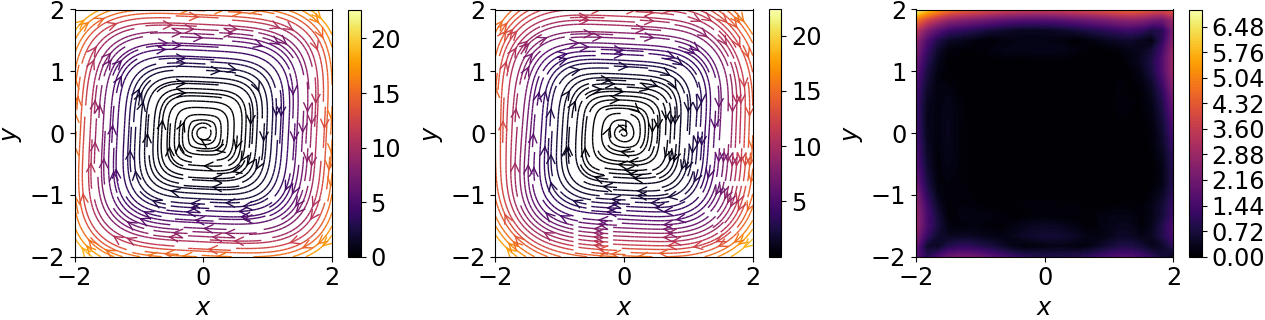}
	\includegraphics[width = 0.39\textwidth, height = 2.2cm,  trim = {11cm 0cm 0cm 0cm }, clip ]{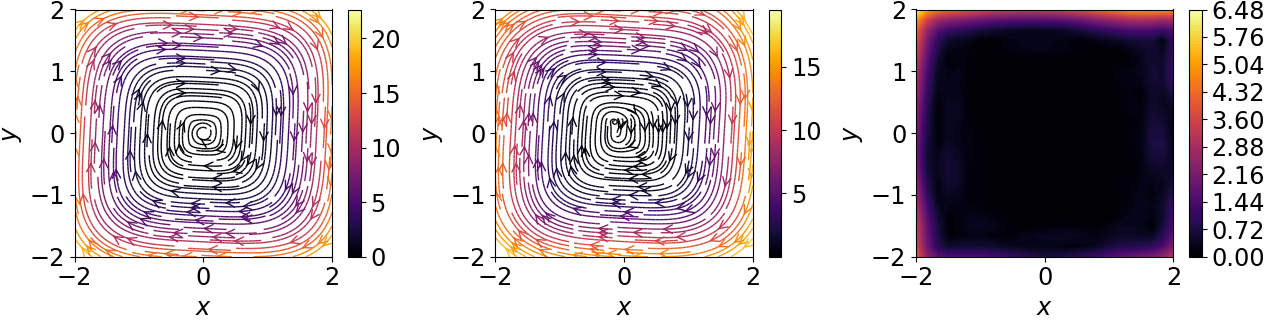} 	\rotatebox{90}{~~~~ \tiny \bf Noise $1\%$}
	
	\includegraphics[width = 0.57\textwidth,height = 2.2cm, ]{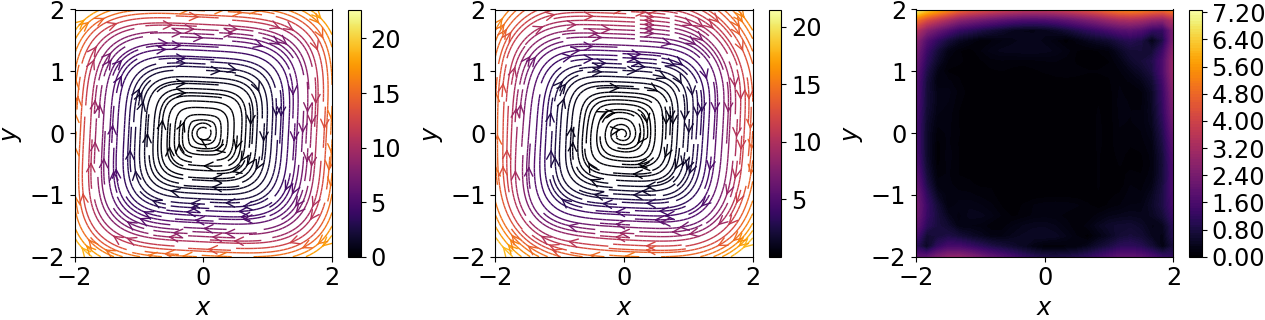}
	\includegraphics[width = 0.39\textwidth, height = 2.2cm,  trim = {11cm 0cm 0cm 0cm }, clip ]{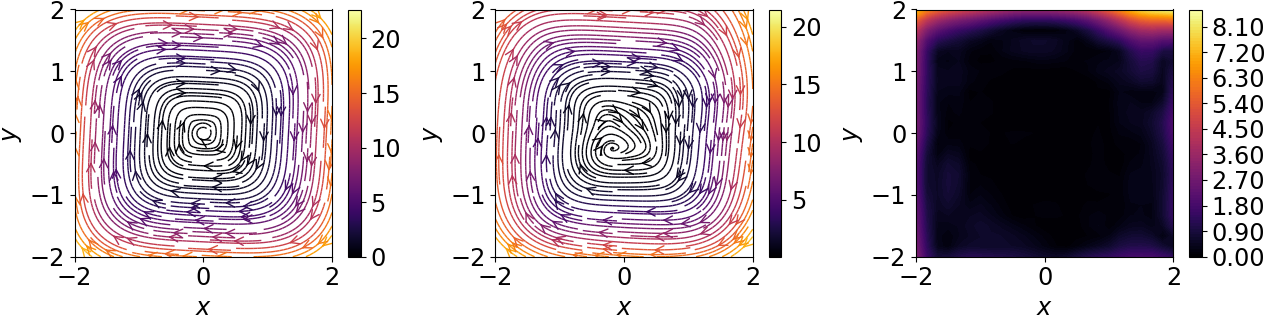} 	\rotatebox{90}{~~~~ \tiny \bf Noise $5\%$}
	
	\includegraphics[width = 0.57\textwidth,height = 2.2cm, ]{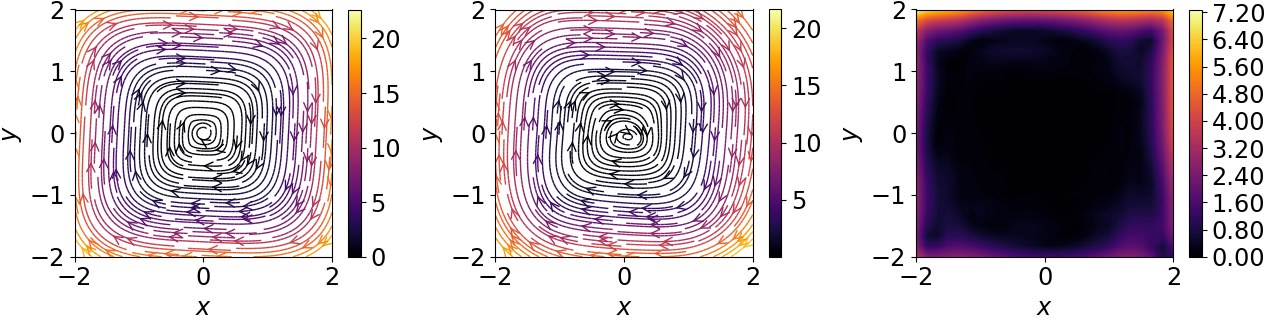}
	\includegraphics[width = 0.39\textwidth, height = 2.2cm,  trim = {11cm 0cm 0cm 0cm }, clip ]{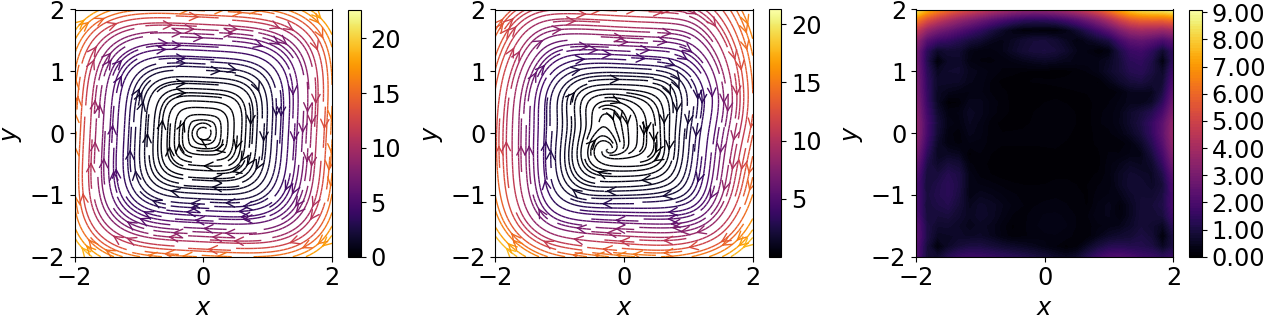} 	\rotatebox{90}{~~~~ \tiny \bf Noise $10\%$}
	
	\includegraphics[width = 0.57\textwidth,height = 2.2cm, ]{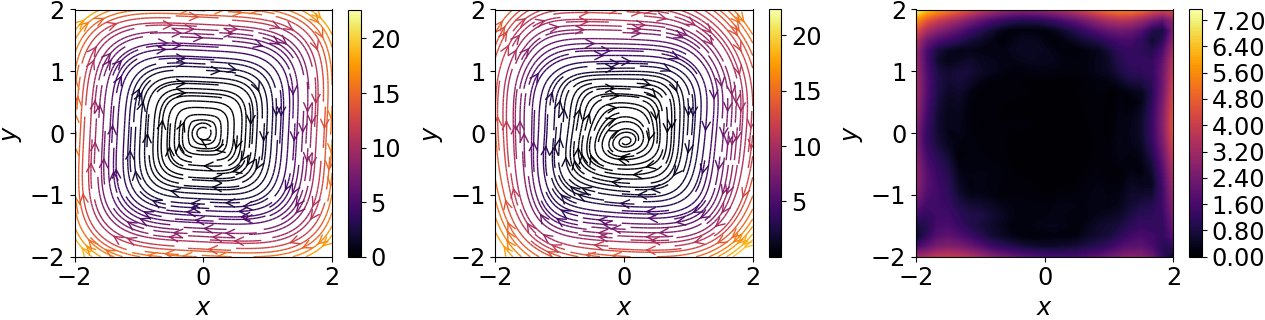}
	\includegraphics[width = 0.39\textwidth, height = 2.2cm,  trim = {11cm 0cm 0cm 0cm }, clip ]{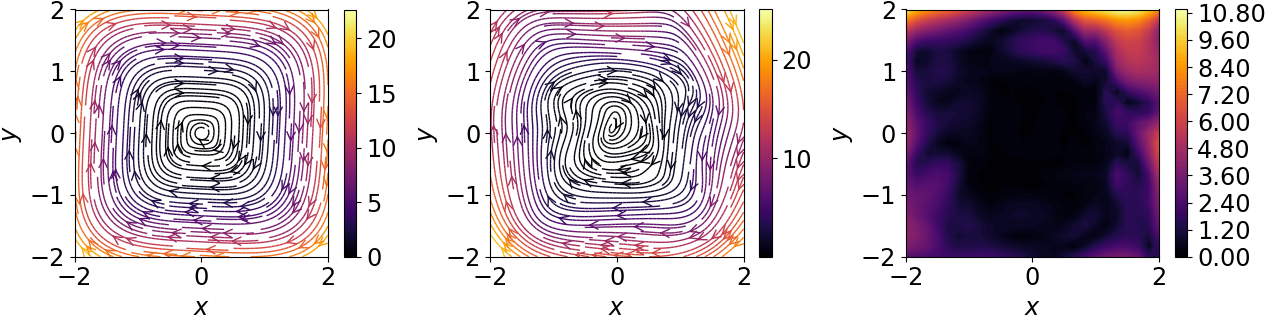} 	\rotatebox{90}{~~~~ \tiny \bf Noise $20\%$}
	
	\includegraphics[width = 0.57\textwidth,height = 2.2cm, ]{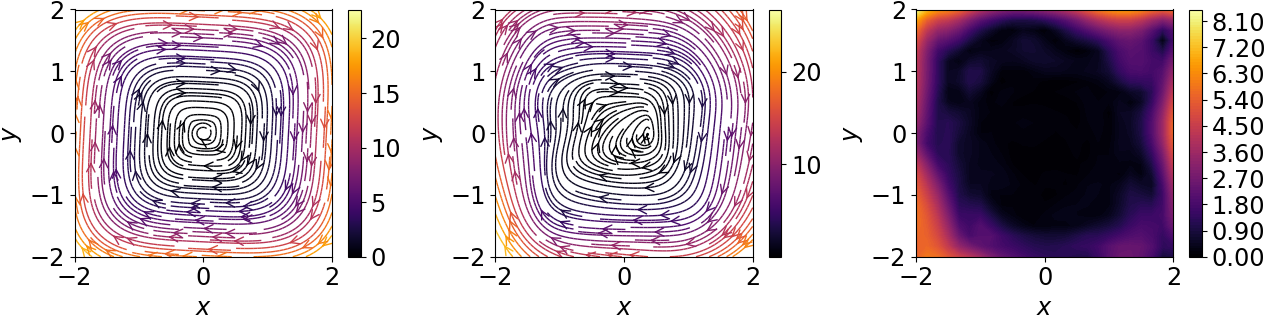}
	\includegraphics[width = 0.39\textwidth, height = 2.2cm,  trim = {11cm 0cm 0cm 0cm }, clip ]{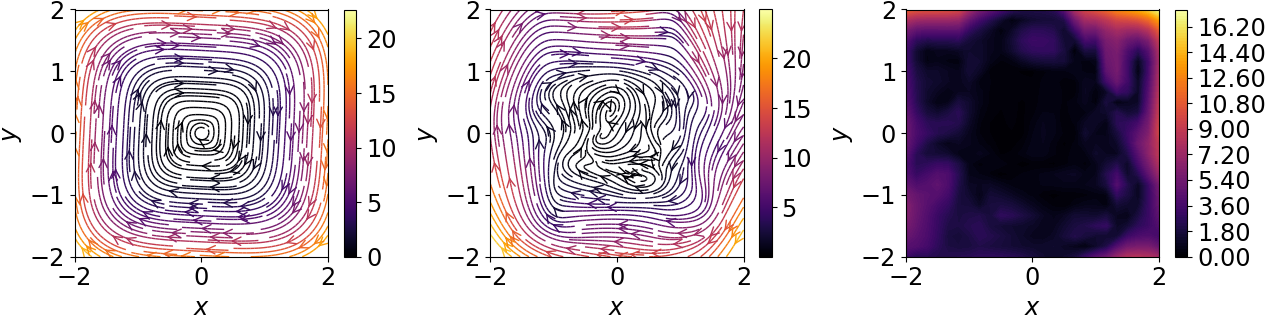} 	\rotatebox{90}{~~~~ \tiny \bf Noise $30\%$}
	
	\includegraphics[width = 0.57\textwidth,height = 2.2cm, ]{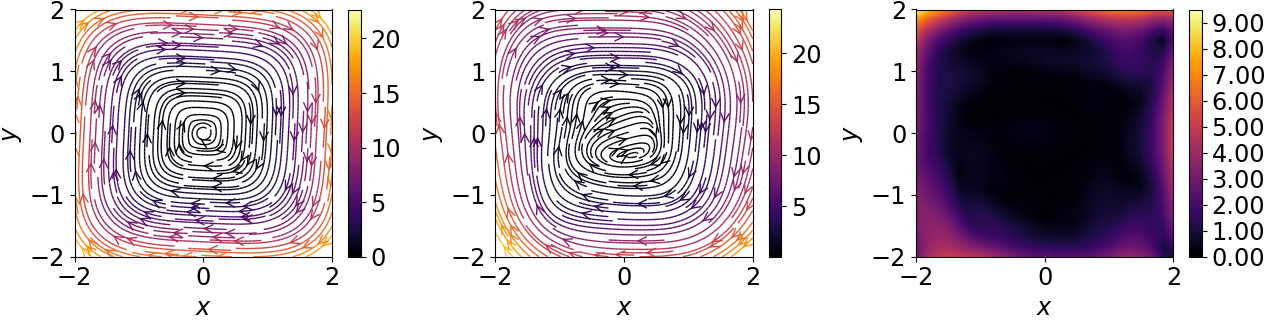}
	\includegraphics[width = 0.39\textwidth, height = 2.2cm,  trim = {11cm 0cm 0cm 0cm }, clip ]{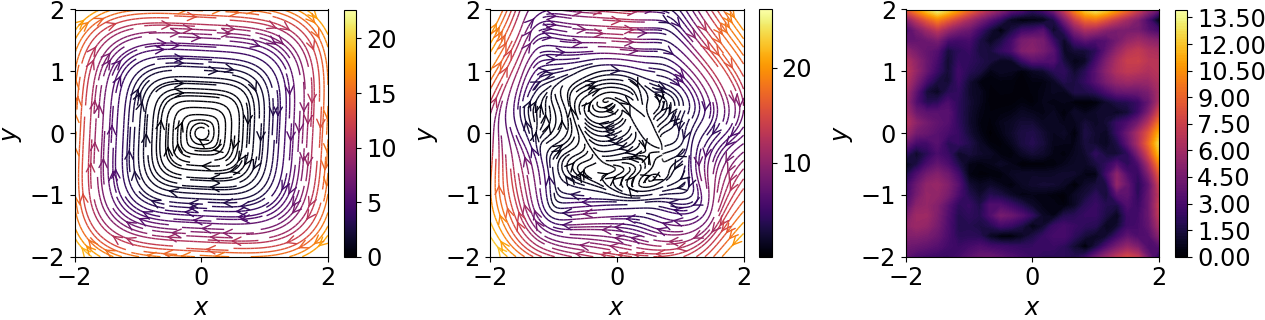} 	\rotatebox{90}{~~~~ \tiny \bf Noise $40\%$}
	\caption{Cubic2D example: A comparison of vector fields of the ground-truth and learned models for various noise. The first column shows the ground truth vector field; the second and fourth columns show the learned vector fields from the \inode~and \snode, respectively; the third and fifth columns present the errors between the learned and the ground truth vector fields.}
	\label{fig:cubic2D_VF}
\end{figure}
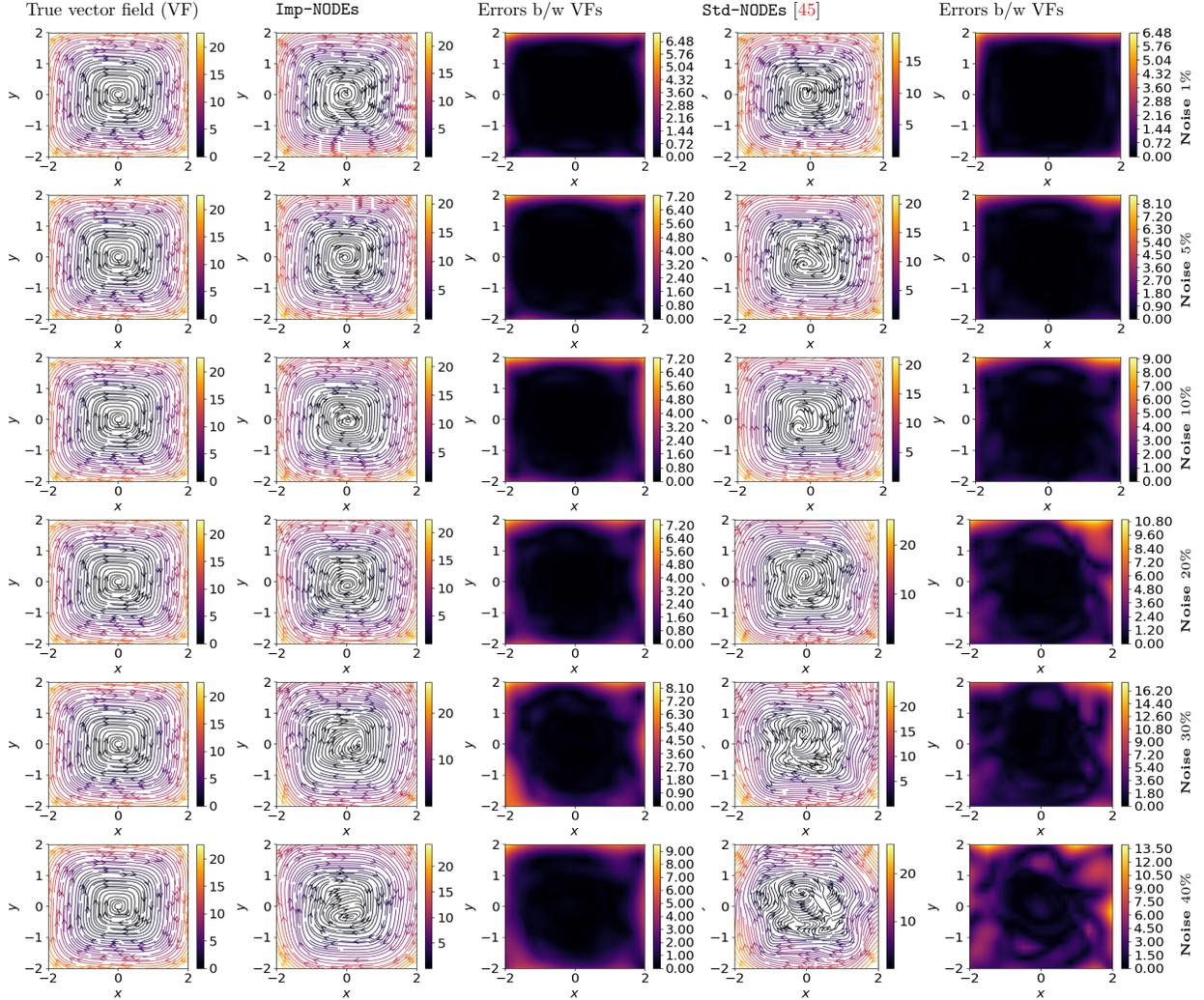

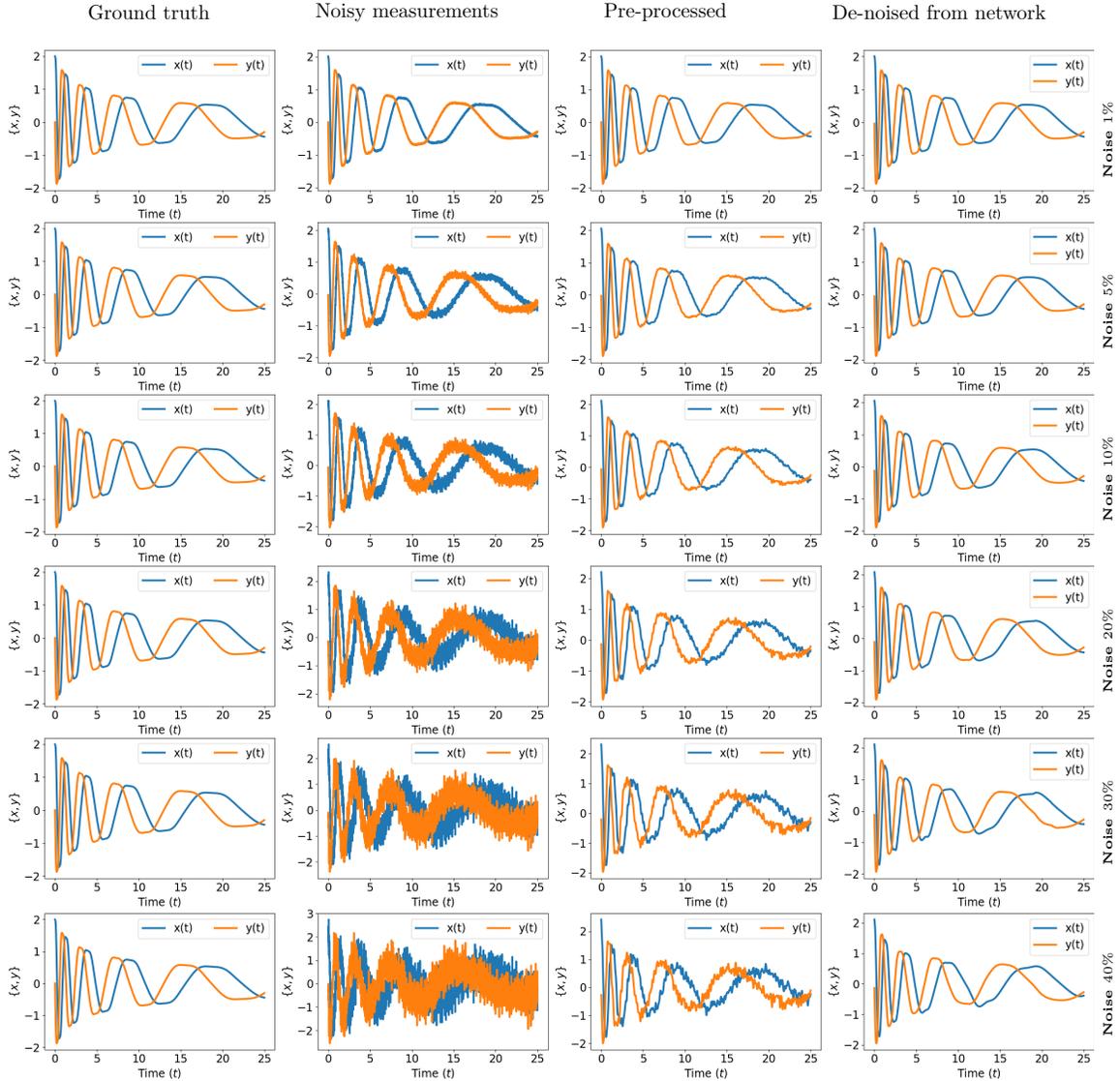
\begin{figure}[!tb]
	\begin{tikzpicture}[scale=0.8, every node/.style={scale=0.8}]
		\node[] (l1) {};
		\node[right = 1.5cm of l1] (l2) {Ground truth};
		\node[right = 1.250cm of l2] (l3) {Noisy measurements};
		\node[right = 1.25cm of l3] (l4) {Pre-processed };
		\node[right = 1.25cm of l4] (l5) {De-noised from network};
	\end{tikzpicture}
	\begin{center}
		\includegraphics[width = 0.9\textwidth, trim = 0cm 0cm 0cm 0cm, clip]{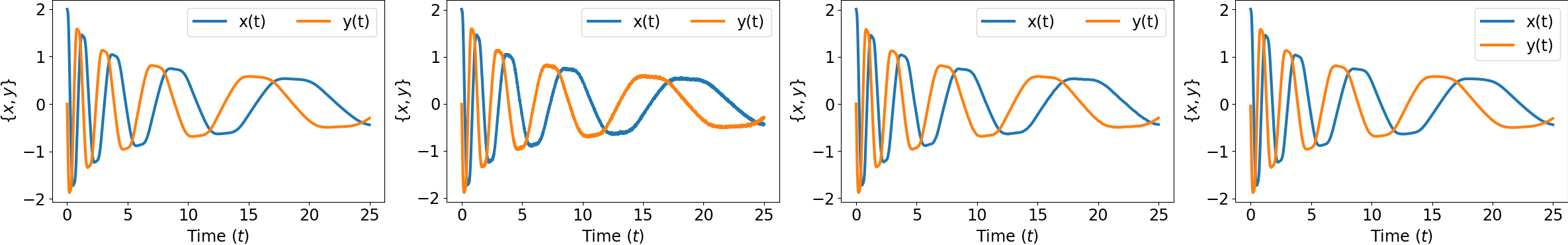}~\rotatebox{90}{~~~~ \tiny \bf Noise $1\%$}
		
		\includegraphics[width = 0.9\textwidth, trim = 0cm 0cm 0cm 0cm, clip]{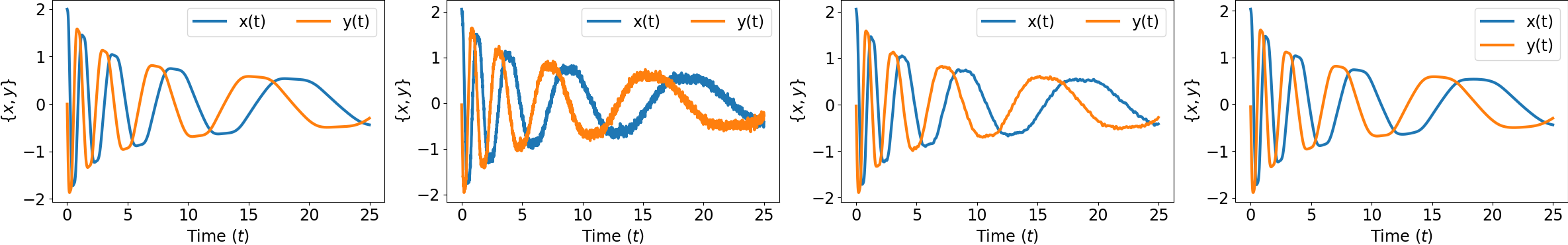}~\rotatebox{90}{~~~~ \tiny \bf Noise $5\%$}
		
		\includegraphics[width = 0.9\textwidth, trim = 0cm 0cm 0cm 0cm, clip]{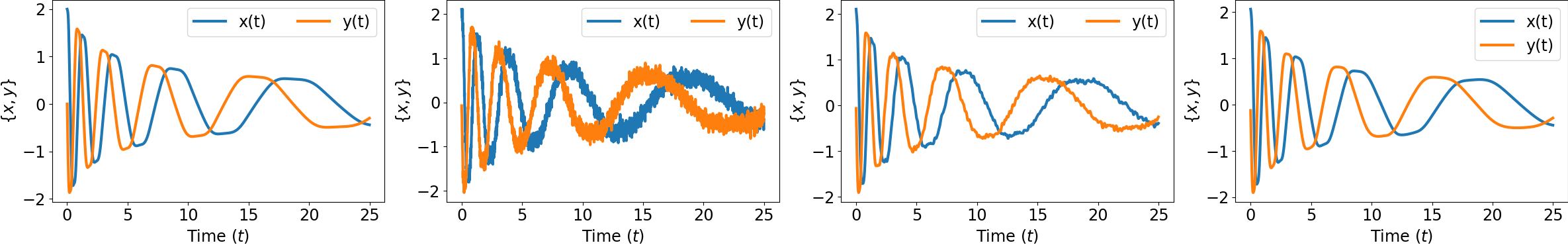}~\rotatebox{90}{~~~~ \tiny \bf Noise $10\%$}
		
		\includegraphics[width = 0.9\textwidth, trim = 0cm 0cm 0cm 0cm, clip]{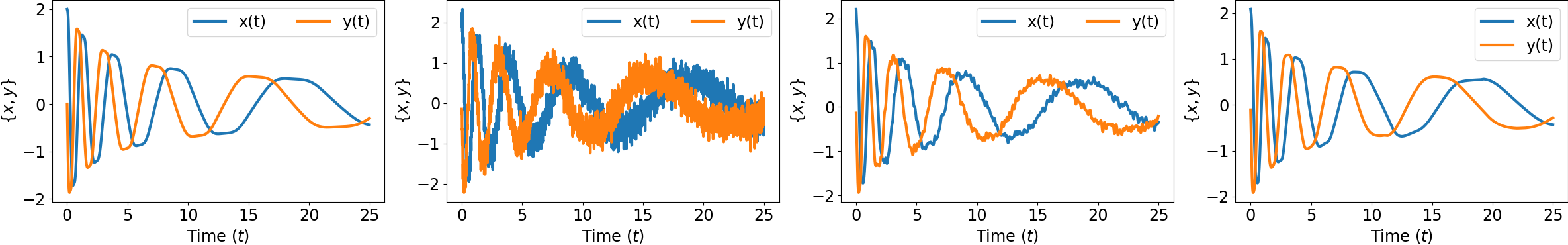}~\rotatebox{90}{~~~~ \tiny \bf Noise $20\%$}
		
		\includegraphics[width = 0.9\textwidth, trim = 0cm 0cm 0cm 0cm, clip]{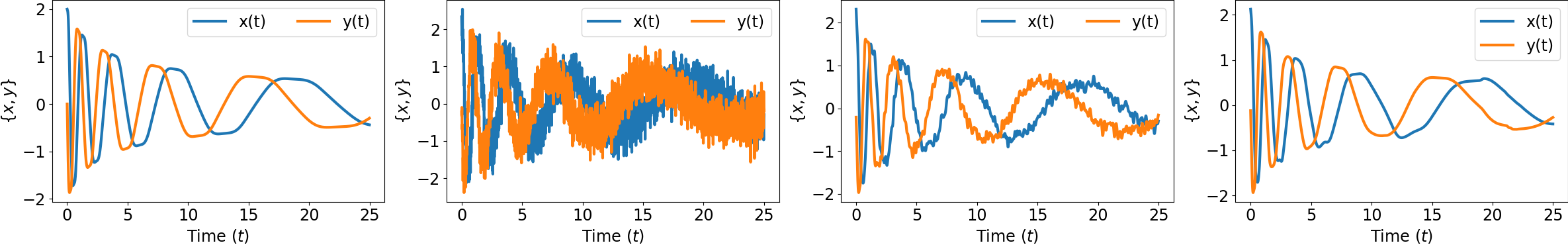}~\rotatebox{90}{~~~~ \tiny \bf Noise $30\%$}
		
		\includegraphics[width = 0.9\textwidth, trim = 0cm 0cm 0cm 0cm, clip]{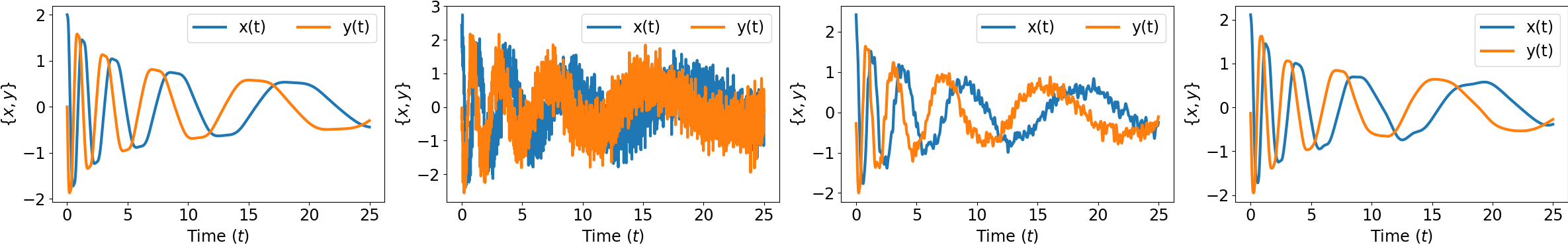}~\rotatebox{90}{~~~~ \tiny \bf Noise $40\%$}
	\end{center}
	
	\caption{Cubic2D example: Denoised data from the trained implicit network for various noise levels. The plots contain the ground truth, noisy measurements, pre-processed data using a low-pass filter, and denoised data from the implicit network.}
	\label{fig:cubic2D_denoised}
\end{figure}

Having trained models, we plot the results in \Cref{fig:cubic2D_VF}, where the learned vector fields from the proposed method (\inode), and it is compared with neural ODE \cite{chen2018neural} (\snode). \snode~is also trained with the same configurations and the same number of epochs as \inode.
It is clear from the figures that \inode~is able to learn the underlying vector field faithfully, whereas the \snode~fails to identify the vector field accurately. It is quite evident for higher levels of noise. Our approach consists of an implicit network, aiming to generate denoised data whose dynamics can be defined by a neural network. Thus, we plot the denoised data obtained from the implicit network in \Cref{fig:cubic2D_denoised}. It shows that using the implicit network, we can obtain denoised data, close to the ground truth clean data even for a high noise level, which, otherwise, is not possible by employing solely \snode.

\begin{figure}[!tb]
	\centering
	\begin{subfigure}[t]{0.45\textwidth}
		\centering
		\includegraphics[width =1\textwidth]{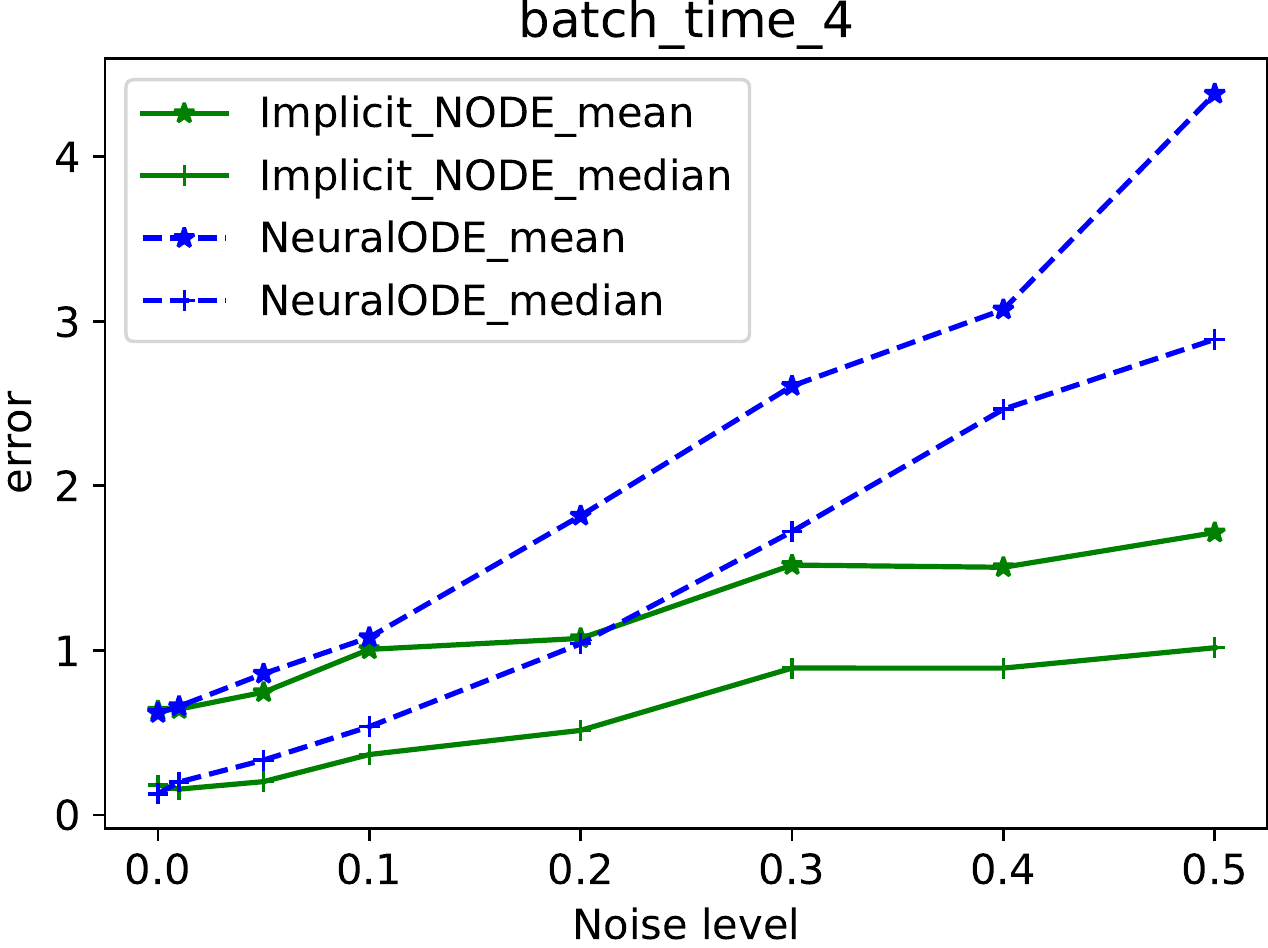}~~~~
		\caption{Mean and median error of the vector fields for \inode~and \snode~for various levels of noise.}
		\label{fig:cubic2D_noise}
	\end{subfigure}\hfill
	\begin{subfigure}[t]{0.45\textwidth}
		\centering
		\includegraphics[width = 1\textwidth]{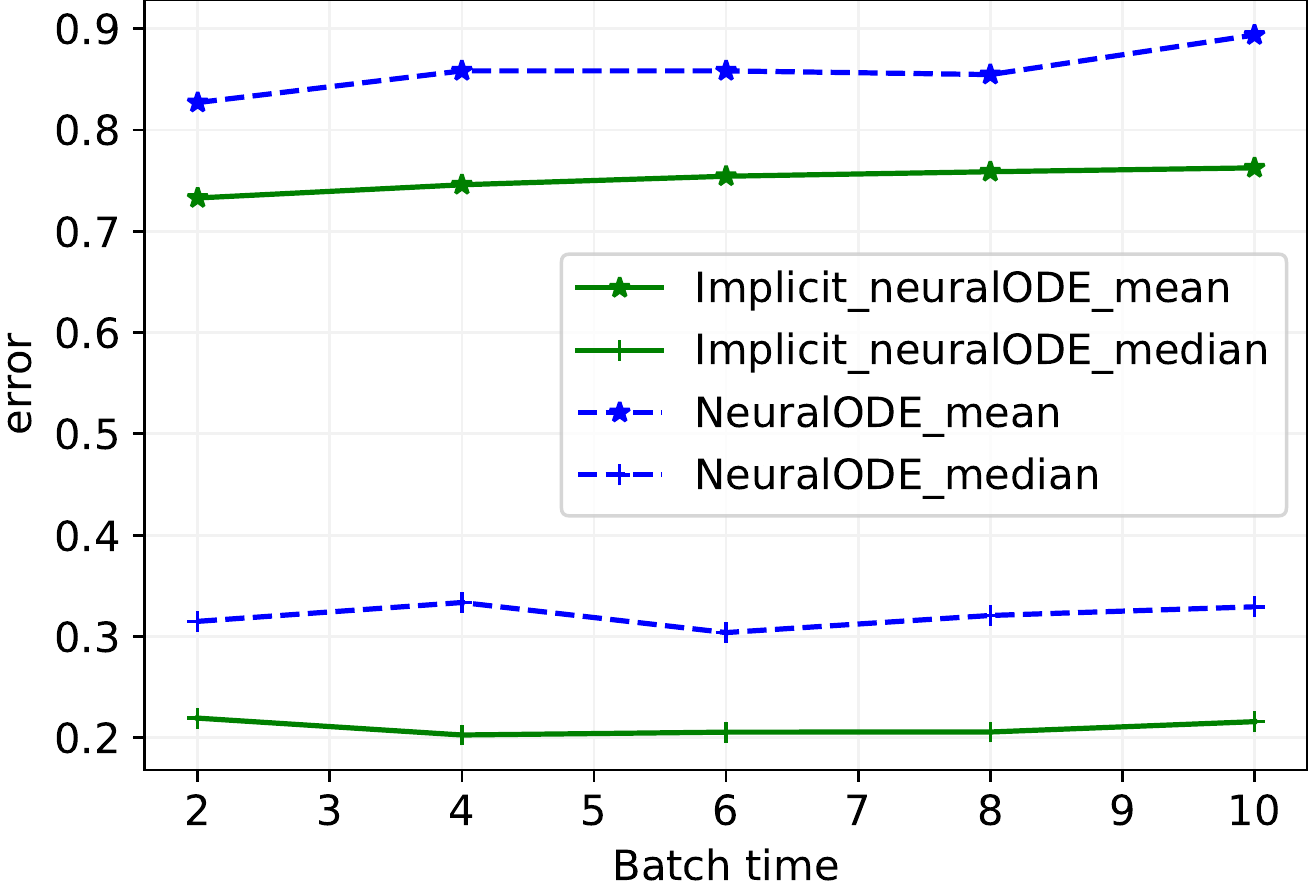}
		\caption{Mean and median error of the vector fields for \inode~and \snode~for different batch size.}
		\label{fig:cubic2D_bs}
	\end{subfigure}
	\caption{Cubic2D example: Comparisons of the error between the learned and ground truth vector fields for different noise levels and for different batch sizes. }
\end{figure}

We further compute the mean and median errors between the learned vector fields and the ground truth for a quantitative comparison of these approaches to learning vector fields. For batch size $\texttt{bs} = 4$, we plot these errors in \Cref{fig:cubic2D_noise}, showing the robustness of the proposed approach with respect to the noise levels. Furthermore, we report the performance of both approaches by changing the batch size, meaning by varying the time span for integration (for batch size $\texttt{bs}$, the time span is $\texttt{bs}\times \dt$) for the corrupted data with $5\%$ noise. For this study, we again plot the mean and medium errors between the learned and ground truth vector fields in \Cref{fig:cubic2D_bs}. We notice that the errors are of the same order with respect to the batch size for \inode~and perform better than  \snode~in both mean and median errors for all batch sizes. We highlight that although the errors are of the same order with respect to \texttt{bs}, we notice a substantial increase in computational cost because of a large integration time span. 

Based on these studies, we empirically conclude that an implicit network looks at generating data in the vicinity of the measurements, which can be integrated by an ODE defined by a neural network. 

\section{Second Order Neural ODEs for Noisy Measurements}\label{sec:secondOrder}
Several dynamics of engineering process, particularly in electrical and mechanical ones are of second-order, which can be given as follows:
\begin{equation}\label{eq:SO_model}
 \ddot{\bx}(t) = \mathbf{f}(\bx(t),\dot{\bx}(t)),
\end{equation}
where $\dot{\bx}(t)$ and $\ddot{\bx}(t)$ denote the first and second derivatives of $\bx(t)$ with respect to $t$, respectively. As discussed in \cite{norcliffe2020_sonode}, it is advantageous to consider the companion first-order system of \eqref{eq:SO_model} which is 
\begin{equation}\label{eq:SO_model_firstorder}
 \begin{bmatrix}  \ddot{\bx}(t) \\ \dot{\bx}(t)  \end{bmatrix}
= \begin{bmatrix}  \mathbf{f}(\bx(t),\dot{\bx}(t)) \\ \dot{\bx}(t) \end{bmatrix},
\end{equation}
which inherently preserves the second-order behavior. 
The above system can be seen as a first-order system with a constraint. The method proposed in the previous section can be readily applied to learn second-order neural ODEs for noisy measurements by incorporating implicit networks. 

\subsection*{Numerical example: Pendulum dynamics}
To illustrate learning second-order dynamics, we consider a  nonlinear pendulum model described as
\begin{equation}
	\ddot{\bx}(t) = -\sin(\bx(t)) - 0.05\cdot \dot{\bx}(t).
\end{equation}
We collect data using the initial condition $[\dot{\bx}(t),\bx(t)] = [-0.5, 2.0]$ at the time interval $\dt = 0.25$s, which is then corrupted by adding Gaussian white noise of $\mu = \{5\%, 20\%\}$. We do not apply any pre-processing step for this example since filtering with a large time-step is not trivial using a simple filter such as a low-pass filter. Hence, we  will also observe the filtering capability of the implicit network with the raw data to obtain denoised data.  We employ the proposed scheme by combining an implicit network and  neural ODEs by imposing the second-order structure. We train networks with parameters  $\lambda_{\integral} = 1.0$, $\lambda_{\grad} = 10^{-2}$, and $\lambda_{\mse} = 1.0$ in \eqref{eq:loss_func} for $\mu = 5\%$, and $\lambda_{\mse} = 0.1$ for $\mu = 20\%$ to avoid over-fitting. We also use  an early-stopping for over-fitting, as discussed in the previous example. For integration, we take the time-span of $2\cdot \dt$.

\begin{figure}[!tb]
 	
 	\begin{tikzpicture}[scale=0.8, every node/.style={scale=0.8}]
 		\node[] (l1) {};
 		\node[right = 0.10cm of l1] (l2) {True vector field (VF)};
 		\node[right = 0.30cm of l2] (l3) {\inode};
 		\node[right = 1.00cm of l3] (l4) {Error b/w VFs};
 		\node[right = 1.25cm of l4] (l5) {\texttt{SO}-\snode~\cite{norcliffe2020_sonode}};
 		\node[right = 0.75cm of l5] (l6) {Error b/w VFs};
 	\end{tikzpicture}
	
	\includegraphics[width = 0.57\textwidth,height = 2.2cm, ]{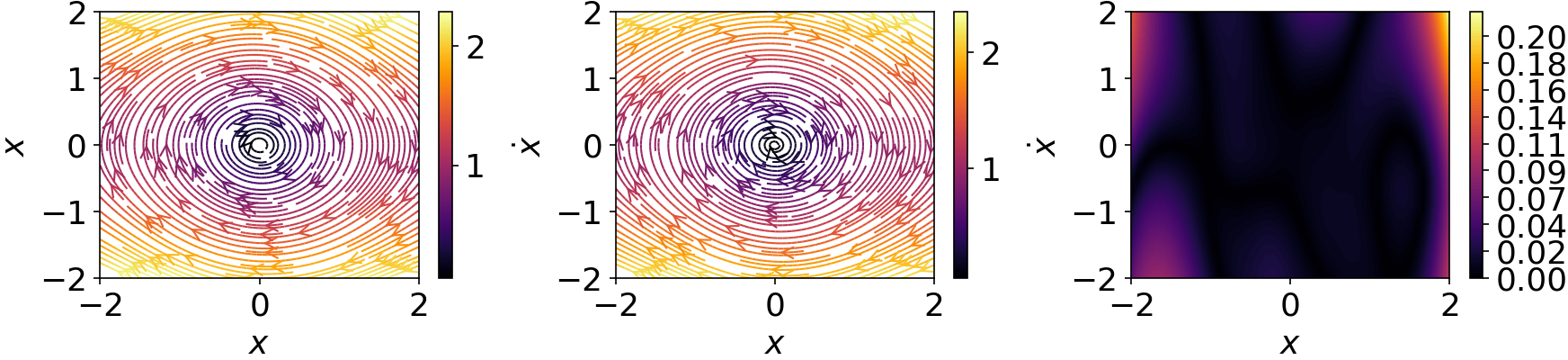}
	\includegraphics[width = 0.39\textwidth, height = 2.2cm,  trim = {9.5cm 0cm 0cm 0cm }, clip ]{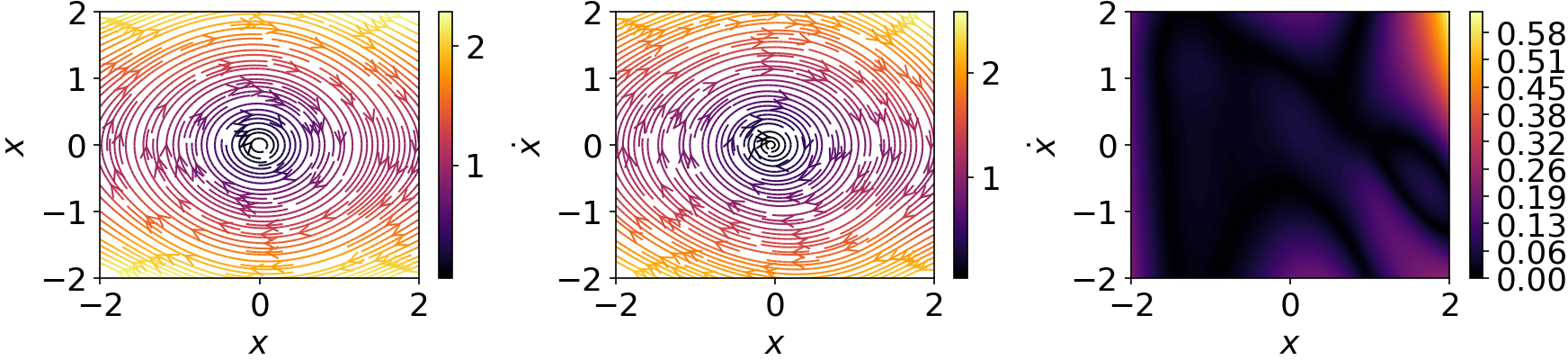} 	\rotatebox{90}{~~~~ \small \bf Noise $5\%$}

	\includegraphics[width = 0.57\textwidth,height = 2.2cm, ]{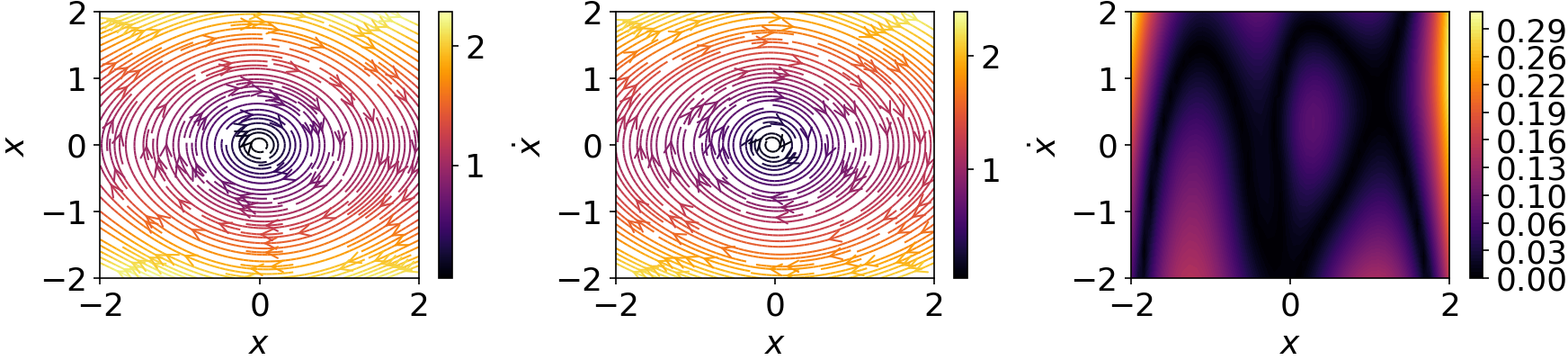}
	\includegraphics[width = 0.39\textwidth, height = 2.2cm,  trim = {9.5cm 0cm 0cm 0cm }, clip ]{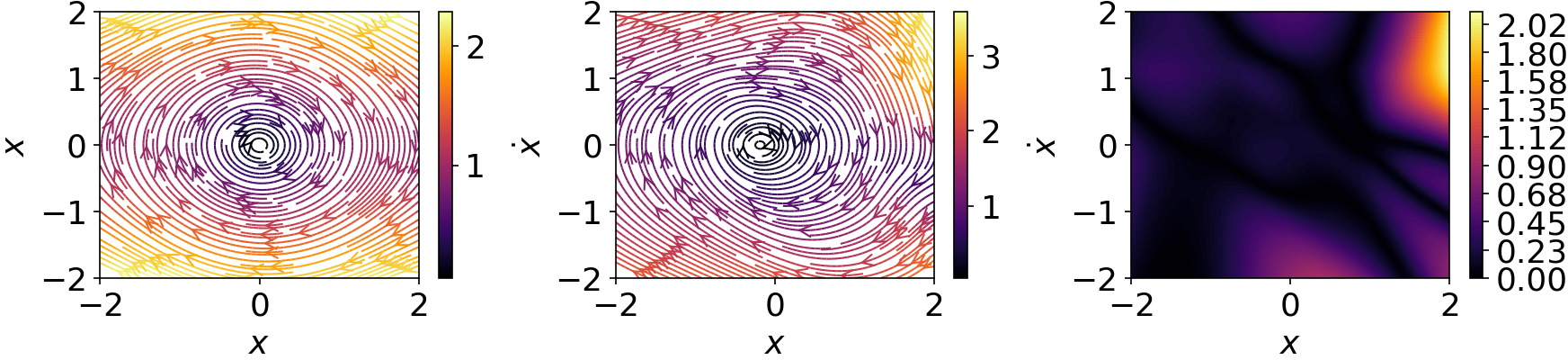} 	\rotatebox{90}{~~~~ \small \bf Noise $20\%$}
	
	\caption{Pendulum example: A comparison of the learned vector fields for second-order dynamical models using the proposed methodology and the one proposed in \cite{norcliffe2020_sonode} for noise $\{5\%,20\%\}$. The leftmost figure shows the ground truth vector field; the second and fourth columns are obtained using the proposed methodology and the one in \cite{norcliffe2020_sonode}, respectively; and the third and fifth columns are the corresponding errors by comparing them with the ground truth.}
\label{fig:pendulum_so}
\end{figure}
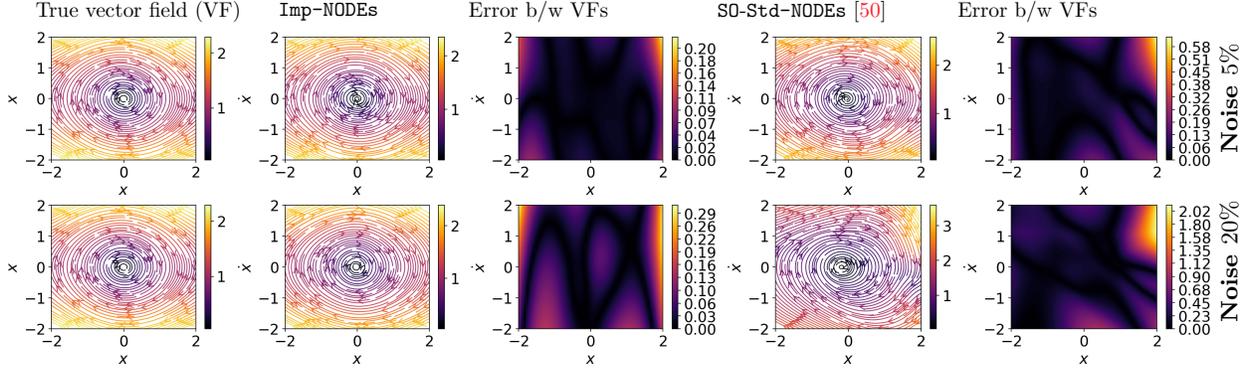

\begin{figure}[!tb]
	\centering
	\includegraphics[width = 0.4\textwidth]{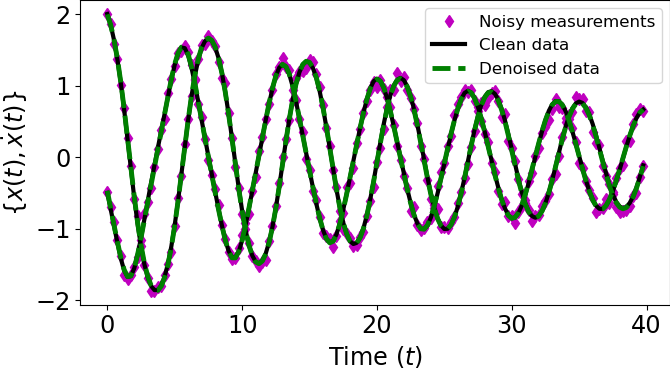} \rotatebox{90}{~~~~\qquad \tiny \bf Noise $5\%$} \qquad
	\includegraphics[width = 0.4\textwidth ]{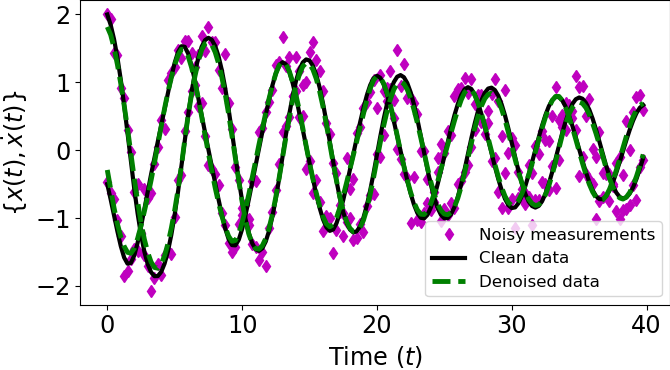}\rotatebox{90}{~~~~\qquad \tiny \bf Noise $20\%$}
	\caption{Pendulum example: The figures show the ground truth, noisy measurements, and denoised data obtained from the implicit network. }
	\label{fig:pendulum_denoised}
\end{figure}

We compare our results with neural ODEs for second-order systems, the approach proposed in \cite{norcliffe2020_sonode}. We plot the learned vector field from both methods in \Cref{fig:pendulum_so}, where we see better performance for the proposed method than the one proposed in \cite{norcliffe2020_sonode}. Particularly, it is quite apparent for a larger noise, see \Cref{fig:pendulum_so}, where \snode~for second-order systems fails to compare the vector field (second row). Moreover, we also plot the denoised data, which is the output of the trained implicit network in \Cref{fig:pendulum_denoised}, indicating recovery of the data faithfully without using any prior pre-processing step as such. 

\section{Measurements at Irregular Sampling}\label{sec:irregularsampling}
Lastly, we illustrate the ready applicability of the proposed method when the data are collected at an irregular time grid, especially when dependent variables are not even measured in the same time frame. It is of particular interest in medical applications, where often data comes at quite irregular time intervals or when the sources of information are different. A similar problem has been discussed in \cite{rubanova2019latent}, but for this, an initial condition is estimated using irregular trajectory points using ODE-RNN. It is then followed by integrating neural ODEs. Assessing the initial condition from the given irregular points is quite challenging, especially when the sequence is long, and also integrating a long sequence imposes additional challenges. Moreover,  although the data can be collected at irregular intervals, all dependent variables still need to be measured/estimated on the same time grid.
On the other hand, we can readily apply our proposed methods when all dependent variables are collected on different time-grids because of learning an implicit representation of the data. 

We here present the framework for two-dimensional problems; however, it readily extends arbitrary dimensional dynamics. Let us consider a dynamical model as:
\begin{equation}
	\dot{\bx}(t) = \mathbf{f}(\bx(t)),
\end{equation}
where $\bx = [\bx_1,\bx_2] \in \R^2$. Next, assume that the variable $\bx_1$ is measured at the time-grid $T_1 = \{t^{(1)}_1,\ldots,t^{(1)}_n\}$, whereas the variable $\bx_2$ is collected at the time-grid $T_2 = \{t^{(2)}_1,\ldots,t^{(2)}_m\}$ with $T_1\neq T_2$. To learn a model for the vector field representing the dynamics for $\bx$ using measurements at an irregular time-grid, we construct an implicit representation for $\bx$ so that both variables can estimated on the same time-grid (let us denote it by $T = \{t_1,\ldots,t_p\}$) but with a constraint using measurements. Assume the implicit network and neural ODE defining the vector field are denoted by $\cN_{\theta}^{\textsf{I}}$ and $\cN_{\phi}^{\textsf{Dyn}}$. To train the network, we define the following loss function:

\begin{equation*}
	\begin{aligned}
		&\lambda_{\mse} \left(\left\|\left[\cN_{\theta}^{\textsf{I}}(t^{(1)}_i)\right]_{1} - \bx_1(t^{(1)}_i)\right\| +  \left\|\left[\cN_{\theta}^{\textsf{I}}(t^{(1)}_i)\right]_{2} - \bx_1(t^{(2)}_i)\right\| \right)
		+  \lambda_{\grad}\left\| \cN_{\phi}^{\textsf{Dyn}}(\cN_{\theta}^{\textsf{I}}(t_i)) -  \dfrac{d}{dt}\cN_{\theta}^{\textsf{I}}(t_i) \right\| \\ &+ \lambda_{\integral}\left\| \left(\cN_{\theta}^{\textsf{I}}(t_j) - \cN_{\theta}^{\textsf{I}}(t_i) \right) - \int_{t_i}^{t_j}\cN_{\phi}^{\textsf{Dyn}}(\cN_{\theta}^{\textsf{I}}(\tau))\d\tau \right\|,
	\end{aligned}
\end{equation*}
where $\left[\cdot\right]_k$ denotes its $k$-element.  

\subsection*{Numerical example: Linear 2D}
\begin{figure}[tb!]
 	\begin{tikzpicture}[scale=0.8, every node/.style={scale=0.8}]
		\node[] (l1) {};
		\node[right = 4.75cm of l1] (l2) {True vector field (VF)};
		\node[right = 1.00cm of l2] (l3) {\inode};
		\node[right = 1.50cm of l3] (l4) {Error b/w VFs};
	\end{tikzpicture}
	
	\vspace{-0.25cm}
	\includegraphics[width = 0.26\textwidth]{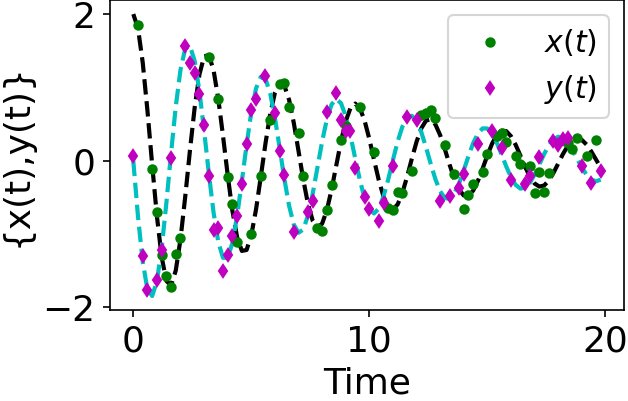}\hfill
	\includegraphics[width = 0.68\textwidth]{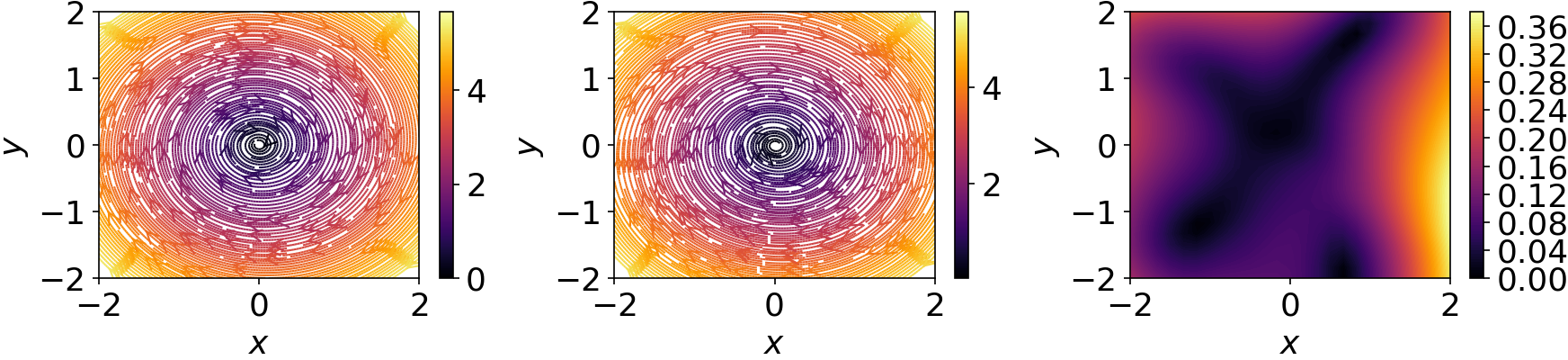} ~~\rotatebox{90}{~~~~  \bf Noise $10\%$}
	\includegraphics[width = 0.26\textwidth]{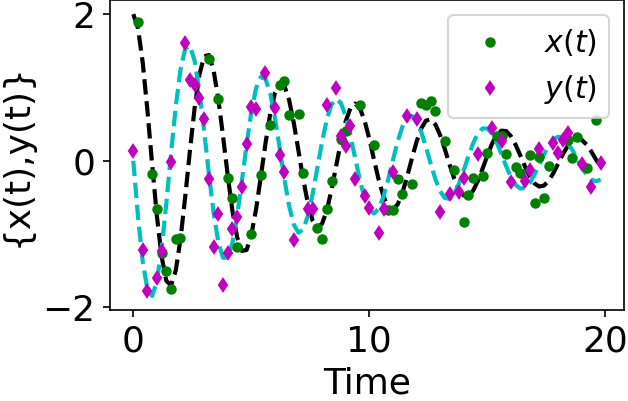}\hfill
	\includegraphics[width = 0.68\textwidth]{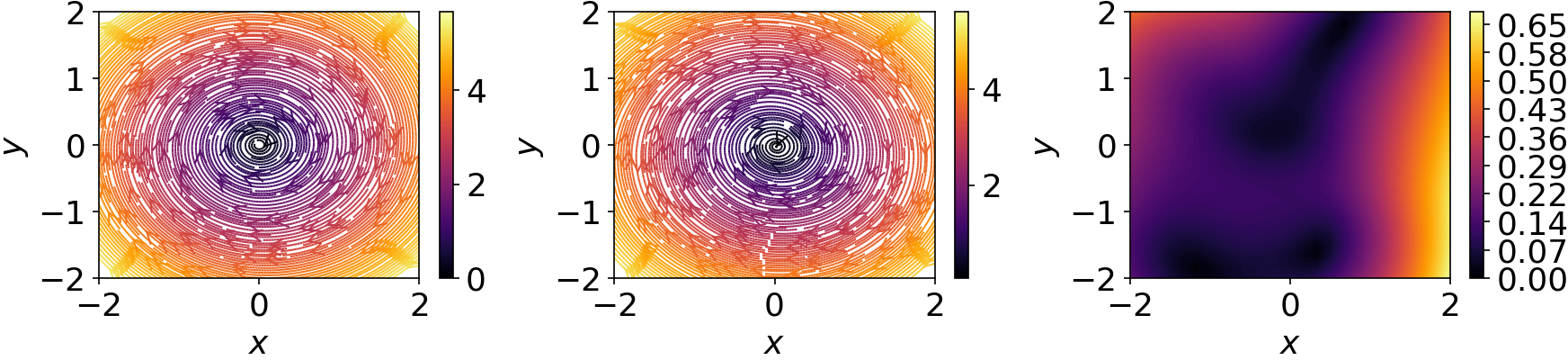} ~~\rotatebox{90}{~~~~  \bf Noise $20\%$}
	\caption{Linear 2D example: 
	The figures illustrate the ground truth and learned vector field using \inode. The left-most plots show the irregularities in the sampled measurements, which clearly shows that the $\bx$ and $\by$ do not share the same time grid. The dotted lines in the left-most figure show the ground truth full trajectories of $\bx$ and $\by$.}
	\label{fig:Linear2D_irregular}
\end{figure}
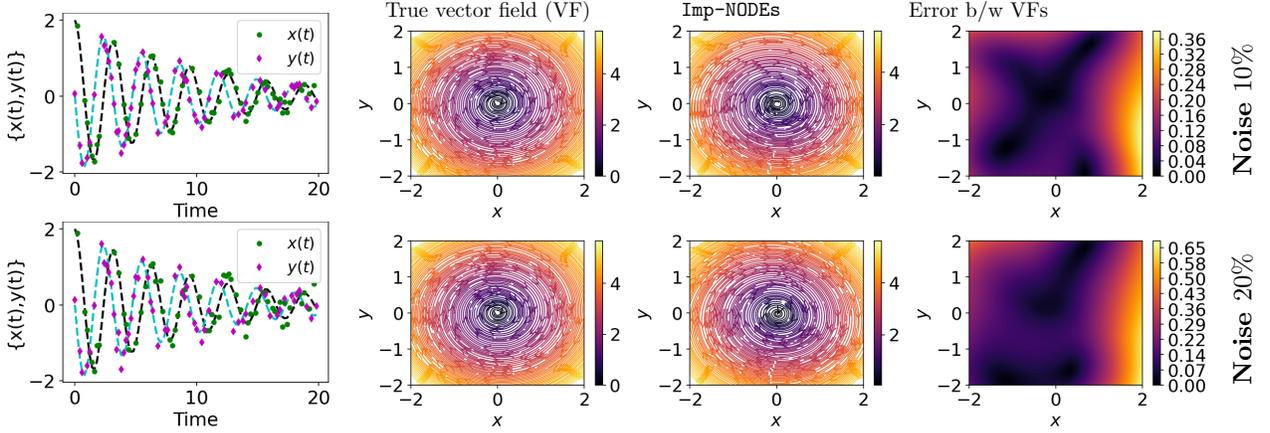

We illustrate the considered scenario using a linear 2D example, given by
\begin{equation*}
	\begin{aligned}
		\dot{\bx}(t) &= 0.1\cdot \bx(t)  +2.0\cdot \by(t),\\
		\dot{\by}(t) &=  2.0\cdot\bx(t)  -0.1\cdot \by(t).
	\end{aligned}
\end{equation*}
We collect data using an initial condition $[\bx,\by] = [2,0]$ with a time interval $\dt = 0.2$ in the time interval $[0,20]$. We  randomly collect $60\%$ independently for the first and second dependent variable, followed by corrupting them using Gaussian white noise of $\mu = \{10\%, 20\%\}$, as shown in the left-most column of \Cref{fig:Linear2D_irregular}. Also, we take the time-grid for prediction of the output of the implicit network as the uniform grid with $\dt = 0.2$ for the time-interval $ [0,20]$.
For learning models for the vector field, we set $\lambda_{\integral} = 1.0$, $\lambda_{\grad} = 10^{-2}$, and $\lambda_{\mse} = 1.0$ for $\mu = 10\%$ and $\lambda_{\mse} = 0.1$ for $\mu = 20\%$. We also do early-stopping for $\mu = 20\%$. For time integration, we consider the time span $\dt$. 

We show the estimates of the learned vector fields using the proposed methodology and are compared with the ground truth in \Cref{fig:Linear2D_irregular},  illustrating faithful capturing the dynamics.  Moreover, we can also recover the clean signal faithfully without any prior information about the noise and any pre-processing of the data. 

\section{Discussion and Conclusion}
This work has presented a new approach for learning dynamical models from highly noisy time-series data and obtaining denoise data. Our framework blends universal approximation capabilities of deep neural networks with neural ODEs.
The proposed scheme involves two networks to learn (approximately)  an implicit representation of the measurement data and of the vector field. 
These networks are combined by enforcing that an ODE can explain the dynamics of the output of the implicit network. 
We also discussed its extension to second-order neural ODEs to learn second-order dynamical models using corrupted data.
Furthermore, we have presented that the proposed approach can readily handle arbitrary sampled points in time. The dependent variables need not be collected at the same time-grid.  It is because we first construct an implicit representation of the data that does not require data to be at a particular grid. 

In the future, we focus on utilizing the encoder-decoder framework combined with an implicit network to learn latent ODEs to explain even richer dynamics of the measurement data. Moreover, when the data are high-dimensional (e.g., coming from partial-differential equations), applying neural ODEs becomes computationally intractable. However, it is known that the dynamics often lie in a low-dimensional manifold. Therefore, in our future work, we aim to utilize the concept of low-dimensional embedding to make learning computationally more efficient for high-dimensional data. Furthermore, it would be interesting to make use of expert knowledge and physical law to have physics-obey neural ODEs so that the generalizability and extrapolation capabilities of models can be further improved. Moreover, the performance of the proposed approach depends on the hyper-parameters that weigh different terms in the loss function. Therefore, we focus on developing an automatic mechanism to determine these parameters.

\bibliographystyle{ieeetr}
\bibliography{mybib}

\appendix

\section{Suitable Architectures and Chosen Hyper-parameter}\label{appendix:models}
\begin{figure}[!tb]
	\centering
	\includegraphics[width = 0.9\textwidth]{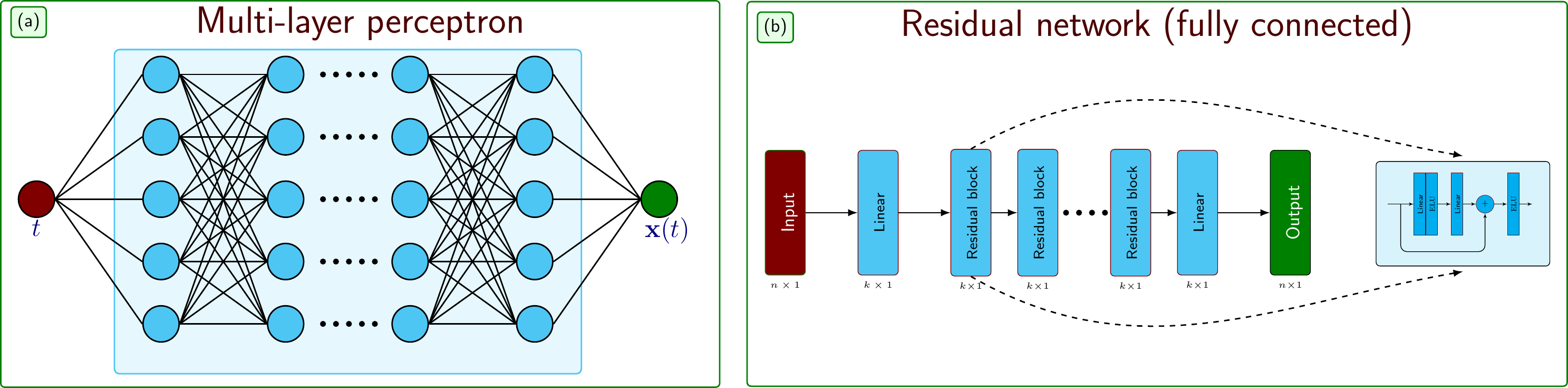}
	\caption{The figure shows three potential simple architectures that can be used to learn either implicit representation or to approximate the underlying vector field. Diagram (a) is a simple multi-layer perceptron, and (b) is a residual-type network but fully connected. }
	\label{fig:NN_archi}
\end{figure}

Here, we briefly discuss neural network architectures suitable for our proposed approach. We require two neural networks for our framework, one for learning the implicit representation $\cN_\theta^{\textsf{I}}$ and the second one  $\cN_\theta^{\textsf{Dyn}}$ is to learn the vector field. For implicit representation, we use a fully connected multi-layer perceptron (MLP) as depicted in \Cref{fig:NN_archi}(a) with periodic activation functions (e.g., $\sin$) \cite{sitzmann2020implicit} which has shown its ability to capture finely detailed features as well as the gradients of a function. To approximate the vector field, we consider a simple residual-type network as illustrated in \Cref{fig:NN_archi}(b) with \emph{exponential linear unit} (ELU) as an activation function \cite{clevert2015fast}. We choose ELU as the activation function since it is continuous and differentiable and resembles a widely used activation function, namely rectified linear unit (ReLU). Furthermore, to train implicit networks, we map the input data to  $[-1,1]$ as recommended in \cite{sitzmann2020implicit}. For the considered examples in the paper, we report the architecture designs (e.g., numbers of neural and layers) in \Cref{tab:NN_info_ODE}.

\begin{table}[!tb]
	\begin{tabular}{|c|c|c|c|c|}\hline 
		Example                           & Networks                       & Neurons & \begin{tabular}[c]{@{}c@{}}Layers or\\ residual blocks\end{tabular} & Learning rates \\ \hline 
		\multirow{2}{*}{Cubic oscillator} & For implicit representation   & 20      & 4                                                                   &         $ 10^{-3}$\\
		& For approximating vector field & 20      & 4                                                                   &     $ 10^{-3}$   \\ \hline 
		
			\multirow{2}{*}{Pendulum} & For implicit representation   & 32      & 4                                                                   &         $5\cdot 10^{-4}$\\
		& For approximating vector field & 20      & 4                                                                   &     $ 10^{-3}$   \\ \hline 
			\multirow{2}{*}{Linear example} & For implicit representation   & 20      & 3                                                                  &         $5\cdot 10^{-4}$\\
		& For approximating vector field & 20      & 2                                                                  &     $ 10^{-3}$   \\ \hline 
	\end{tabular}
	\caption{The table shows the information about network architectures and learning rates.}
	\label{tab:NN_info_ODE}
\end{table}

\end{document}